# Real-time object detection method for embedded devices

ZICONG JIANG[1], LIQUAN ZHAO[1], SHUAIYANG LI[1] and YANFEI JIA[2]
[1]Key Laboratory of Modern Power System Simulation and Control and Renewable Energy Technology, Ministry of Education, Northeast Electric Power University, jilin, jilin 132012, China
[2]College of Electrical and Information Engineering, Beihua University, Jilin 132013, China;

Corresponding author: Zhao Liquan (: zhao_liquan@163.com; Tel.: +86-150-4320-1901).

This work was supported by the National Natural Science Foundation of China (61271115)

**ABSTRACT** The "You only look once v4" (YOLOv4) is one type of object detection methods in deep learning. YOLOv4-tiny is proposed based on YOLOv4 to simple the network structure and reduce parameters, which makes it be suitable for developing on the mobile and embedded devices. To improve the real-time of object detection, a fast object detection method is proposed based on YOLOv4-tiny. It firstly uses two ResBlock-D modules in ResNet-D network instead of two CSPBlock modules in Yolov4-tiny, which reduces the computation complexity. Secondly, it designs an auxiliary residual network block to extract more feature information of object to reduce detection error. In the design of auxiliary network, two consecutive 3x3 convolutions are used to obtain 5x5 receptive fields to extract global features, and channel attention and spatial attention are also used to extract more effective information. In the end, it merges the auxiliary network and backbone network to construct the whole network structure of improved YOLOv4-tiny. Simulation results show that the proposed method has faster object detection than YOLOv4-tiny and YOLOv3-tiny, and almost the same mean value of average precision as the YOLOv4-tiny. It is more suitable for real-time object detection, especially for developing on embedded devices .

## I. INTRODUCTION

Object detection method based on deep learning mainly includes two types: region proposal-based two-stage method and regression-based one-stage method [1-2]. The typical two- stage methods include region-based convolution neural network (R-CNN) method [3], Fast R-CNN [4], Faster R-CNN [5] method, region-based fully convolutional networks (R-FCN) method [6], light head R-CNN method and other improve method based on convolution neural network [7-8]. Although two-stage method has higher accuracy than one-stage method, the one-stage method has faster detection speed than two-stage method [9-10]. The one-stage method is more suitable for application in some conditions that require higher real-time.

The You Only Look Once (YOLO) method [11] proposed by Redmon, et al. is the first regression-based one stage method. Redmon, et al. also proposed the You Only Look Once version 2 (YOLOv2) [12] based on YOLO by deleting fully connected layer and the last pooling layer, using anchor boxes to predict bounding boxes and designing a new basic network named DarkNet-19. The You Only Look Once version 3 (YOLOv3) [13] is the last version of YOLO method proposed by Redmon, et al. It introduces feature pyramid network, a batter basic network named darknet-53 and binary cross-entropy loss to improve the detection accuracy and the ability of detecting smaller object. Due to the type of information fusion employed by YOLOv3 does not make full use of low-level information, a weakness which restricts its potential application in industry. Therefore, Peng, et al. have proposed the YOLO-Inception method [14], one which uses the inception structure with diversified receptive fields, which in turn can provide rich semantic information and improve the performance of small object detection. Tian has proposed the YOLOv3-dense method [15]. It uses the dense net method to process feature layers with low resolution, which effectively enhances feature propagation, promotes feature reuse, and improves network performance. Two years later, after the authors of YOLOv3 declared to give up updating it, Alexey, et al. proposed the YOLOv4 method [16] that has been accepted by the authors of YOLOv3. It used CSPDarknet53 backbone, spatial pyramid pooling module, PANet path-aggregation neck and YOLO3 (anchor based) head as the architecture of YOLOv4. Besides, it also introduced a new method of data augmentation mosaic and self-adversarial training, applied genetic algorithms to select optimal hyper-parameters and modified some existing method to make the proposed method suitable for efficient training and detection.

YOLO serial methods and their improved methods have complex network structure and a larger number of network parameters. They require powerful GPU (graphic processing unit) computing power to realize the real-time object detection. However, they have limited computing power and limited memory, and require real-time object detection for some mobile devices and embedded devices (autonomous driving devices, augmented reality devices and other smart device) in real-world applications [17]. For example, such as real-time inference on smart phones and embedded video surveillance, the available computing resources are limited to a combination of low-power embedded GPUs or even just embedded CPUS with limited memory. Therefore, it is a big challenge to realize the real-time object detection on embedded devices and mobile devices. To solve the problem, lightweight object detection methods are proposed by many researchers. The lightweight methods have comparatively simpler network structure and fewer parameters. Therefore, they require lower computing resources and memory, and have faster detection speed. They are more suitable for deploying on mobile devices and embedded devices. Although they have lower detection accuracy, the accuracy can meet the actual demands. Lightweight object detection methods based on deep learning have been applied in many fields, including vehicle detection [18-19], pedestrian detection [20], bus passenger object detection [21], agricultural detection [22], human abnormal behavior detection [23], etc.

A number of lightweight object detection methods have already been proposed to improve detection speed with the limitation of hardware platforms and meanwhile to meet the demand of high performance. Such MobileNet series(MobileNetv1 [24], MobileNetv2 [25], MobileNetv3 [26]), Squeezenet series (Squeezenet [27], SqueezeNext [28]), ShuffleNet series(ShuffleNet_v1 [29], ShuffleNet_v2 [30]) , lightweight YOLO series [31-41]. MobileNet_v1 method [24] constructs lightweight deep neural networks by using depthwise separable convolution instead of the traditional convolution to reduce parameters. Based on MobileNet_v1, the MobileNet_v2 bulids inverted residual module by adding the point-wise convolution layer in front of depthwise separable convolution to improve the ability of extracting features. MobileNet_v3 redesigns some computionally-expensive layers and introduces the hard swish nonlinearity to improve detection speed. Squeezenet method design the new network architecture based on CNN by replacing 3*3 convolutions with 1*1 convolutions, using squeeze layers to decrease the number input channels to 3*3 convolutions and downsampling late in the network to improve detection speed. SqueezeNext method is proposed based on squeezenet. Its neural network architecture is able to achieve AlexNet's top-5 performance with 112 X fewer parameters [28]. The MobileNet series, squeezenet series and shuffleNet series are directly designed to realize lightweight network. The lightweight YOLO series methods are designed based on complete YOLO. They are realized by suppressing the network of complete YOLO method. YOLOv2-tiny is one of lightweight YOLO series methods [31]. The complete YOLOv2 uses the Darknet19 as backbone network, which contains 19 convolution layers and 6 pooling layers. They YOLOv2-tiny method delete convolution layers in Darknet19 network to 9 layers to reduce the network complexity. YOLOv3-tiny is proposed by compressing the network model of YOLOv3 [13]. It uses seven layer convolution networks and six max pooling layers instead of the ResBlock structure in DarkNet53 network [40]. It also reduce the output branch from the three scale predictions(52×52, 26×26 and 13×13) to two scale predictions (26×26 and 13×13). YOLOv4-tiny [41] is also one of lightweight YOLO series methods, and also realized based on YOLOv4 [16]. It uses CSPDarknet53-tiny backbone network instead CSPDarknet53 backbone network of YOLOv4. The spatial pyramid pooling (SPP) and path aggregation network (PANet) are also be instead by feature pyramid networks (FPN) to reduce the detection time. Besides, it also uses two scale predictions (26×26 and 13×13) instead of three scale predictions. Compared with YOLOv3-tiny, the YOLOv4-tiny uses the CSPBlock network to extract feature without using the conditional con convolution networks, and introduces the complete intersection over union to select bounding boxes.

In this section, we have reviewed recent developments rela ted to object detection. In Section 2, we outline the concepts and processes of the YOLOv4-tiny object detection method. I n Section 3 we describe our proposed method. In Section 4, we illustrate and discuss our simulation results.

## II. YOLOv4-tiny

### A. NETWORK STRUCTURE

Yolov4-tiny method is designed based on Yolov4 method to make it have faster speed of object detection. The speed of object detection for Yolov4-tiny can reach 371 Frames per second using 1080Ti GPU with the accuracy that meets the demand of the real application. It greatly increases the feasibility that object detection method is deployed on embedded systems or mobile devices.

The Yolov4-tiny method uses CSPDarknet53-tiny network as backbone network to instead of the CSPDarknet53 network that is used in Yolov4 method. The CSPDarknet53-tiny network uses the CSPBlock module in cross stage partial network instead of the ResBlock module in residual network. The CSPBlock module divides the feature map into two parts, and combines the two parts by cross stage residual edge. This makes the gradient flow can propagate in two different network paths to increase the correlation difference of gradient information. The CSPBlock module can enhance the learning ability of convolution network comparing with ResBlock module. Although this increase computation by 10%-20%, it improves the accuracy. To reduce the amount of calculation, it removes the computational bottlenecks that have higher amount of calculation in CSPBlock module. It improves the accuracy of Yolov4-tiny method in the case of constant or even reduced computation.

To further more simply computation process, Yolov4-tiny method uses the LeakyReLU function as activation function in CSPDarknet53-tiny network without using the Mish activation function that used in Yolov4. The LeakyReLU function is:

$$y_i = \begin{cases} x_i & x_i \geq 0 \\ \dfrac{x_i}{a_i} & x_i < 0 \end{cases} \quad (1)$$

where $a_i \in (1, +\infty)$, it is a constant parameters.

In the part of feature fusion, Yolov4-tiny method uses feature pyramid network to extract feature maps with different scales to increase object detection speed, without using the spatial pyramid pooling and path aggregation network that are used in Yolov4 method. At the same time, the Yolov4-tiny uses two different scales feature maps that are 13×13 and 26×26 to predict the detection results. Supposing that the size of input figure is 416×416 and feature classification is 80, the Yolov4-tiny network structure is shown in figure1.

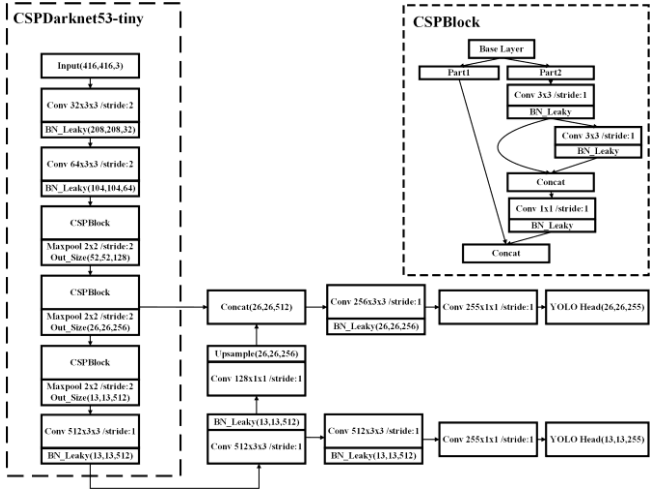

**FIGURE 1.** YOLOv4-tiny network structure.

### B. PREDICTION PROCESS

The prediction process of Yolov4-tiny method is the same with Yolov4 method. It also firstly adjusts the size of input image to make all input images have the same fixed size. Secondly, input images are divided into grids with the size S×S, and every girds will use B bounding boxes to detect object. Therefore, it will generate S×S×B bounding boxes for an input image, and the generated bounding boxes cover the whole input image. If the center of some object falls in some gird, the bounding boxes in the gird will predict the object.

To reduce the redundancy of bounding boxes in prediction process, confidence threshold is proposed. If the confidence score of bounding box is higher than the confidence threshold, the bounding box will be keep; else the bounding box will be deleted. The confidence score of bounding box can be obtained as follows:

$$C_i^j = P_{i,j} * IOU_{pred}^{truth} \quad (2)$$

where $C_i^j$ is the confidence score of the $j$ th bounding box in the $i$ th grid. $P_{i,j}$ is merely a function of the object. If the object is in the $j$ th box of the $i$ th grid, $P_{i,j} = 1$, otherwise $P_{i,j} = 1$. The $IOU_{pred}^{truth}$ represents the intersection over union between the predicted box and ground truth box. The larger the abjectness score, the closer the predicted box is to the ground truth box. The loss function of Yolov4-tiny is the same with Yolov4, which contains three parts. It can be expressed as follows:

$$loss = loss_1 + loss_2 + loss_3 \quad (3)$$

where $loss_1$, $loss_2$ and $loss_3$ are confidence loss function, classification loss function and bounding box regression loss function, respectively. The confidence loss function is that

Another problem with performing domain adaptation using discriminators is that discriminators assign the same importance to different samples; this makes some parts difficult to transfer, which can lead to negative transfer. To solve this problem, the CDAN method applies entropy to the network:

$$loss_1 = -\sum_{i=0}^{S^2}\sum_{j=0}^{B} W_{ij}^{obj}[\hat{C}_i^j \log(C_i^j) + (1-\hat{C}_i^j)\log(1-C_i^j)] - \\ \lambda_{noobj}\sum_{i=0}^{S^2}\sum_{j=0}^{B}(1-W_{ij}^{obj})[\hat{C}_i^j \log(C_i^j) + (1-\hat{C}_i^j)\log(1-C_i^j)] \quad (4)$$

where $S^2$ is the number of grid in input image, $B$ is the number of bounding box in a grid, $W_{ij}^{obj}$ is merely a function of the object. If the $jth$ bounding box of the $ith$ grid is responsible for detecting the current object, $W_{ij}^{obj} = 1$, otherwise $W_{ij}^{obj} = 0$. The $C_i^j$ and $\hat{C}_i^j$ are the confidence score of predicted box and confidence score of truth box, respectively. $\lambda_{noobj}$ is a weight parameter.

The classification loss function is:

$$loss_2 = -\sum_i^{S^2}\sum_j^{B} W_{ij}^{obj}\sum_{c=1}^{C}[\hat{p}_i^j(c)\log(p_i^j(c)) - (1-\hat{p}_i^j(c))\log(1-p_i^j(c))] \quad (5)$$

where $p_i^j(c)$ and $\hat{p}_i^j(c)$ are predict probability and truth probability to which the object belongs to $c$ classification in the $jth$ bounding box of the $ith$ grid.

The bounding box regression loss function is:

$$loss_3 = 1 - IOU + \frac{\rho^2(b, b^{gt})}{c^2}$$
$$+ \frac{16}{\pi^4} \frac{\left(\arctan\frac{w^{gt}}{h^{gt}} - \arctan\frac{w}{h}\right)^4}{1 - IOU + \frac{4}{\pi^2}\left(\arctan\frac{w^{gt}}{h^{gt}} - \arctan\frac{w}{h}\right)^2} \quad (6)$$

where $IOU$ is intersection over union between the boxes that are predicted bounding box and truth bounding box. $w^{gt}$ and $h^{gt}$ are the truth width and height of the bounding box, respectively. $w$ and $h$ are the predicted width and height of the bounding box, respectively. $\rho^2(b, b^{gt})$ denotes the Euclidean distance between the center points of predicted bounding box and truth bounding box. $c$ is the minimum diagonal distance of box that can contain the predicted bounding box and truth bounding box.

## III. PROPOSED METHOD

The Yolov4-tiny method uses the CSPBlock module as residual module. It improves the accuracy, but it also increases network complexity. This reduces the speed of object detection. To improve the speed of object detection with slight impacting accuracy, an improved Yolov4-tiny is proposed.

To speed up the object detection, we use the ResBlock-D module instead of two CSPBlock modules in Yolov4-tiny. The CSPBlock and ResBlock-D modules are shown in figure 2. In ResBlock-D module, it directly uses two paths network to deal with the input feature map. It also has two paths network. The path A network contain three layers that are 1x1 convolutions, 3x3 convolutions with 2 strides and 1x1 convolutions. The path B network contains two layers that are 2x2 average poolings with 2 strides and 1x1 convolutions. Compared with CSPBlock module, the ResBlock-D module deletes the first layer with 3x3 convolutions in CSPBlock, and uses the 1x1 convolutions layer to instead the 3x3 convolutions layer in CSPBlock module in path A to further more reduce computation. Although it increases two layers in path B, the increased computation is smaller than reduced computation. To analysis the computation of two modules, the floating point operations (FLOPs) are used to compute computation complexity. It can be expressed as follow:

$$FLOPs = \sum_{l=1}^{D} M_l^2 \cdot K_l^2 \cdot C_{l-1} \cdot C_l \quad (7)$$

where $D$ is the sum of all convolution layers, $M_l^2$ is output feature map size in $l$th convolution layer, $K_l^2$ is the number of kernel size, $C_{l-1}$ and $C_l$ are the number of input channel and output channel, respectively. We suppose the size of input image is $104 \times 104$ and the number of channel is 64. Based on (7), the FLOPs of CSPBlock used in Yolov4-tiny is:

$$\begin{aligned}FLOPs &= 104^2 \times 3^2 \times 64^2 \\ &+ 104^2 \times 3^2 \times 64 \times 32 \\ &+ 104^2 \times 3^2 \times 32^2 \\ &+ 104^2 \times 1^2 \times 64^2 \\ &= 7.421 \times 10^8\end{aligned} \quad (8)$$

The FLOPs of ResBlock-D used in our proposed method is:

$$\begin{aligned}FLOPs &= 104^2 \times 1^2 \times 64 \times 32 \\ &+ 52^2 \times 3^2 \times 32^2 \\ &+ 52^2 \times 1^2 \times 32 \times 64 \\ &+ 64 \times 52^2 \times 2^2 \\ &+ 52^2 \times 1^2 \times 64^2 \\ &= 6.438 \times 10^7\end{aligned} \quad (9)$$

Based on (8) and (9), the computation complexity rate of CSPBlock and ResBlock-D is about 10:1. It means that the computation complexity of ResBlock-D is smaller than CSPBlock.

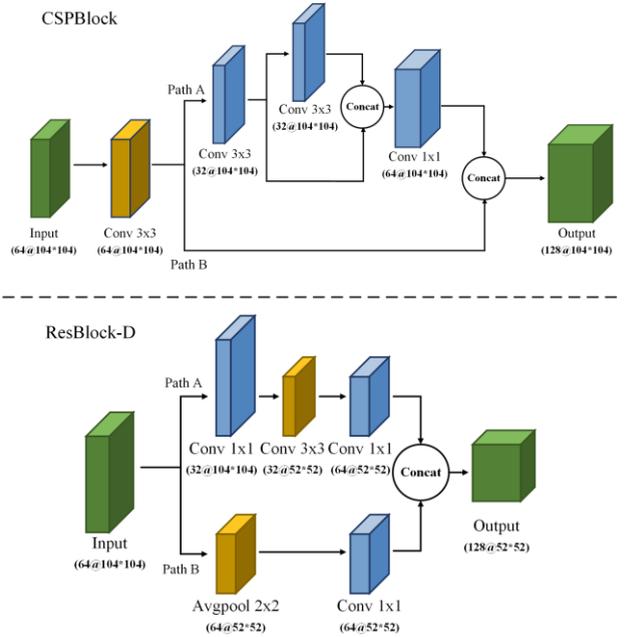

**FIGURE 2.** CSPBlock and ResBlock-D modules.

Although we use the ResBlock-D module to replace CSPBlock module to improve object detection speed, it reduces the accuracy of object detection. To keep the balance between accuracy and speed, we design two same residual network blocks as auxiliary network block and add them into the ResBlock-D module to improve accuracy. The proposed backbone network is shown in figure 3.

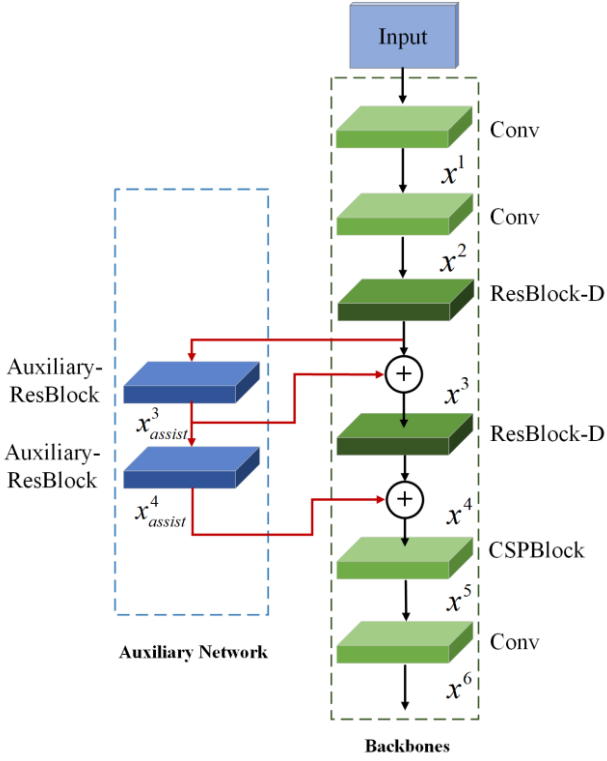

**FIGURE 3. Proposed backbone network.**

The output feature of designed residual network block is fused with shallow feature of backbone network by element-wise summation operation, and the fused information is used as the input of next layer in the backbone network. The fused process can be expressed as:

$$x^l = F^l(x^{l-1}) + x^l_{assist} \quad (l=3,4) \quad (10)$$

where $l$ is the index of network layer, $x^l$ is the output of $l-1$ th layer and also input of $l$ th layer, $x^l_{assist}$ is the output of our designed residual network, $F^l()$ denotes the relationship between the input and output in the $l$ th layer network. This realizes the convergence between deep network and shallow network. It makes the network learn more information to improve detection accuracy and avoids the large increase of calculation.

In original backbone network, the residual network module uses 3x3 convolution kernels to extract feature. The size of its receptive field is also 3x3. Although the smaller receptive field can extract more local information, it losses the global information, which affects the accuracy of object detection. To extract more global feature, we use two consecutive same 3x3 convolutions to obtain 5x5 receptive fields in the auxiliary residual network block. The auxiliary network transmits the extracted global information to backbone network. The backbone network combines the global information obtained by larger receptive field and local information obtained by smaller receptive field to obtain more object information. Besides, with the increased of network depth, the semantic information also becomes more advanced. The attention mechanism can focus on processing and transmitting the effective features, and channel suppresses the invalid features. Therefore, we introduce the channel attention module and spatial attention module into our designed auxiliary network module to extract more effective information. The channel attention module focuses on that 'what' is meaningful given an input image. The spatial attention module focuses on 'where' is an informative part, which is complementary to the channel attention. We directly use the CBAM (Convolutional Block Attention Module) [42] to realize the channel attention and spatial attention simultaneously. The CBAM can be expressed as:

$$\begin{aligned} F' &= M_c(F) \otimes F, \\ F'' &= M_s(F') \otimes F' \end{aligned} \quad (11)$$

where $F \in R^{C \times H \times W}$ denotes input feature map, "$\otimes$" expresses element-wise multiplication, $F''$ is the final output feature map, $M_c()$ and $M_s()$ are channel attention map and spatial attention map, respectively.

The $M_c(F)$ is expressed as:

$$M_c(F) = \sigma(MLP(AvgPool(F)) + MLP(MaxPool(F))) \quad (12)$$

where $AvgPool()$ and $MaxPool()$ denote average-pooling operation and max-Pooling operation, respectively. $MLP()$ denotes Multi-Layer perceptron network, $\sigma()$ is the sigmoid function. The $M_s(F')$ is expressed as:

$$M_s(F') = \sigma(f^{7 \times 7}[MaxPool(F'); AvgPool(F')]) \quad (13)$$

where $f^{7 \times 7}$ is the convolution operation with the kernel size of $7 \times 7$, $[\cdot;\cdot]$ denotes concatenate operation.

Our complete designed auxiliary network block used in figure 3 is shown in figure 4. It uses two $3 \times 3$ convolutions network to extract the global features, and channel attention and spatial attention to extract more effective information. The concatenate operation is used to combine the output feature obtained from the first convolution network and the output feature obtained from the spatial attention. The combined feature is used as the output feature of designed auxiliary network. In the end, the final output feature of auxiliary network will be combined with the output feature of residual network in backbone network to be used as the input feature of next residual network in backbone network. It makes the improved backbone network can extract the global and local feature of detection object, and further more improve the accuracy of detection.

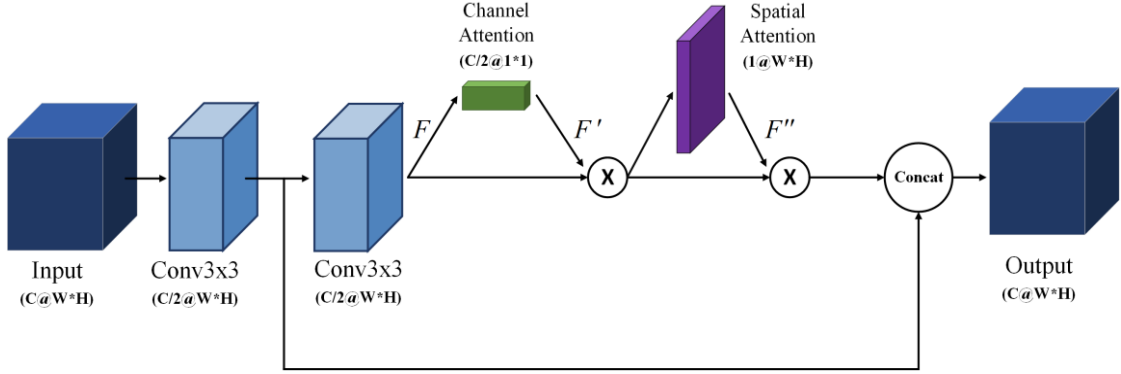

**FIGURE 4.** Auxiliary residual network block.

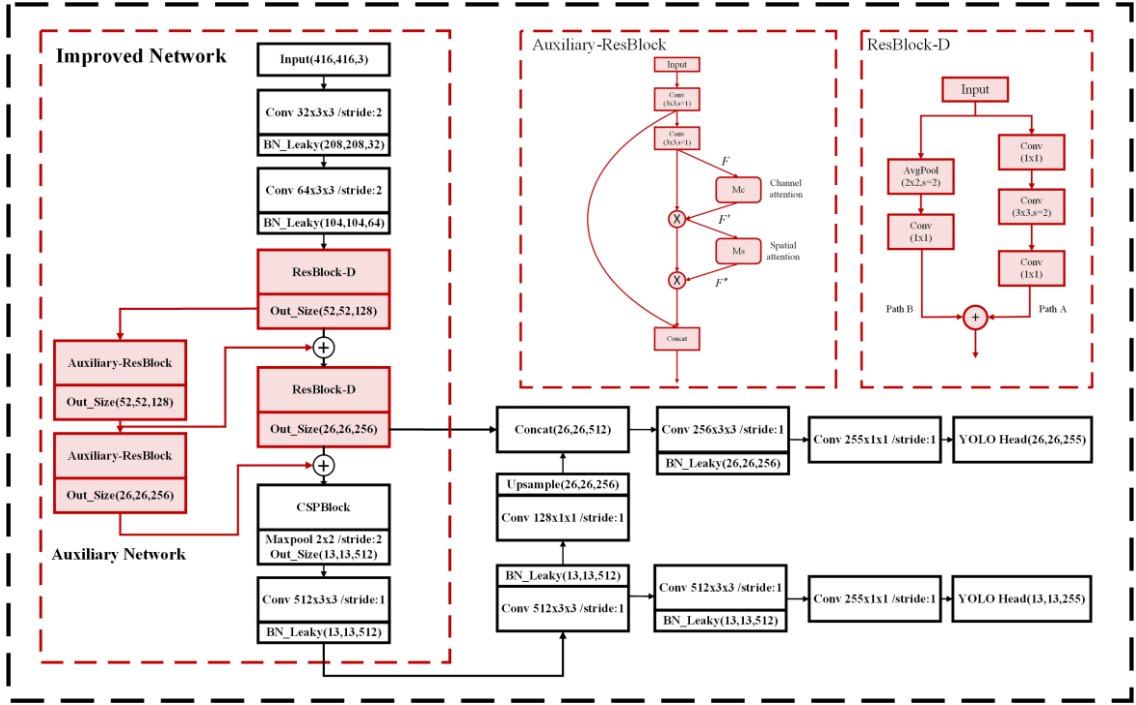

**FIGURE 5.** network structure of our proposed method.

Based on the above introduction, the whole network structure of our proposed YOLO v4-tiny is shown in figure 5. The mainly difference between our proposed method and YOLOv4-tiny in network structure are that we use two ResBlock-D modules to replace two CSPBlock modules in the original CSPNet53-tiny network. Besides, we design auxiliary network block by using two $3 \times 3$ convolutions network, channel attention, spatial attention and concatenate operation to extract global feature. In the end, we merge our designed auxiliary network into the backbone network to build new backbone network. Our proposed network is marked in red in figure 5.

## III. SIMULATION AND DISCUSSIONS

In this paper, we use the MS COCO (Microsoft Common Objects in Context) dataset as train and test dataset. The MS COCO is an authoritative and significant benchmark used in the field of object detection and recognition. It is widely used in many detection methods. It contains 117,264 training images and 5000 testing images with 80 classes. The experiments environment configured in this paper is as follows: The operating system is Ubuntu 18.04. The CPU is Intel Xeon E5-2678 v3 with 2.5 GHZ main frequency. The GPU is NVIDIA GeForce GTX 1080Ti. In order to make full use of the GPU to accelerate the network training, the CUDA 10.1 and its matching CUDNN are installed in the system. The deep learning framework is PyTorch. We use the same parameters for different methods. The batch size, epoch, learning rate, momentum and decay are 16, 273, 0.001, 0.973 and 0.0005 for all methods, respectively.

The mAP (mean value of average precision) , FPS (Frames per second) and GPU utilization are used to quantitatively evaluate the performance of different methods. The mAP is the mean value of average precision for the detection of all classes. FPS denotes the number of images that can be detected successfully in one second. GPU utilization denotes used GPU memory in testing the different detection methods.

We firstly compare our proposed method with YOLOv3、YOLOv4 ,YOLOv3-tiny, YOLOv4-tiny to test their performance in mAP and FPS. The results are shown in Table 1. Although YOLOv4 and YOLOv3 methods have the larger mAP than other methods, they also have the smaller FPS than other methods. YOLOv4 and YOLOv4 methods have complex network structure and many parameters. This makes them have better performance in mAP and worse performance in FPS. They demand the platform to be very powerful. This limits to deploy them on the mobile and embedded devices. YOLOv3-tiny, YOLOv4-tiny and our proposed method belong to lightweight deep learning method. They have relatively simple network structure and few parameters. Therefore, they have better performance in FPS and worse performance in mAP, and more suitable for deploying on the mobile and embedded devices.

Due to the YOLOv3-tiny, YOLOv4-tiny and our proposed methods belong to lightweight deep learning method, and YOLOv3 and YOLOv4 methods do not belong to it, we only compare our proposed method with YOLOv3-tiny and YOLOv4-tiny in the following analysis. Compared our proposed method with YOLOv3-tiny and YOLOv4-tiny, our proposed method has the largest FPS, and YOLOv4-tiny has the largest mAP followed by our proposed method. The mAP of our proposed method is 38% and YOLOv4-tiny method is 38.1%. The relative mAP only reduces by 0.26%. The FPS of our proposed method is 294 and YOLOv4-tiny method is 270. The relative FPS increases by 8.9%. Although the mAP of our proposed method is reduction compared with YOLOv4-tiny, the reduction is much smaller than the increase of FPS, and almost can be ignored.

**TABLE 1. Comparison of different methods in FPS and mAP.**

| Method | FPS | mAP(%) |
| --- | --- | --- |
| YOLOv3 | 49 | 52.5 |
| YOLOv4 | 41 | 64.9 |
| YOLOv3-tiny | 277 | 30.5 |
| **YOLOv4-tiny** | **270** | **38.1** |
| **Proposed method** | **294** | **38.0** |

Table 2 shows the GUP utilization when different methods are used to detect object. GPU utilizations are 1123MB, 1055MB and 1003MB for YOLOv3-tiny, YOLOv4-tiny and proposed method, respectively. The proposed method has the smallest GPU utilization. Based on above analysis, our proposed method has faster detection speed and smaller GPU utilization than others, and is more suitable for developing on the mobile and embedded devices.

**TABLE 2. Comparison of different methods in GPU utilization(MB).**

| Method | GPU utilization(MB) |
| --- | --- |
| YOLOv3-tiny | 1123 |
| YOLOv4-tiny | 1055 |
| Proposed method | 1003 |

We also randomly select six images from testset of MS COCO dataset. The object detection results of YOLOv4-tiny and our proposed method for the six images are shown in figure 6 and figure 7, respectively. From the two figures, we can see that both of the two methods successfully detect the same object. The difference between two figures is the confidence scores for some different object. For sub-figure (a), the confidence scores of train is 0.92 in figure 6 and 0.94 in figure 7. For sub-figure (b), the confidences score of person and surfboard are 0.88 and 0.40 in figure 6 and 0.90 and 0.45 in figure 7, respectively. For figure (c), the confidence scores of three giraffes are 0.84, 0.52 and 0.56 in figure 6, and 0.88, 0.63 and 0.65 in figure 7, respectively. Base on the analysis of the figure (a)-figure (b), all confidence scores obtained by using our proposed method is larger than using YOLOv4-tiny method for the same object.

For sub-figure (d), the confidence scores of three persons and umbrella are 0.58, 0.65, 0.70 and 0.81 in figure 6, and 0.70, 0.74 ,0.67 and 0.81 in figure 7,respectively. Only one confidence score obtained by our proposed method is smaller than YOLOv4-tiniy for sub-figure (d). For figure (e), the confidence scores of four buses are 0.91, 0.91, 0.91 and 0.90 in figure 6, and 0.93, 0.92, 0.93 and 0.81 in figure 7, respectively. Only one bus confidence score obtained by our proposed method is smaller than YOLOv4-tiniy for sub-figure (e). For sub-figure (f), the confidence scores of four persons, three win glasses and one laptop are 0.51, 0.78, 0.75, 0.39 ,0.37, 0.76,0.54 and 0.17 in figure 6, and 0.65, 0.69, 0.85 ,0.54, 0.37, 0.53,0.32 and 0.22 in figure 7. The confidence scores for four objects (three persons and one laptop) obtained by our proposed method are larger than obtained by YOLOv4-tiny method, and one confidence score is the same for two methods. Although the number of detection object is 22 in six images, there are only five confidence scores obtained by YOLOv4-tiny are larger than obtained by our proposed method. Based on the above analysis, our proposed method has better performance for detecting larger object, and larger confidence score than YOLOv4-tiny method for most objects, when both of two methods successfully detect object.

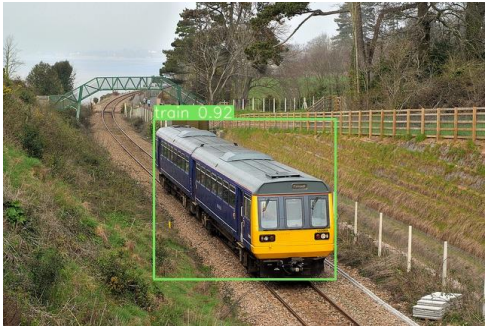
(a)
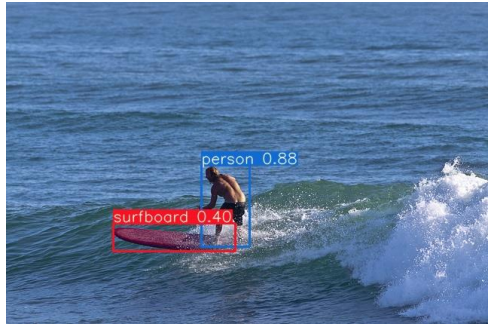
(b)
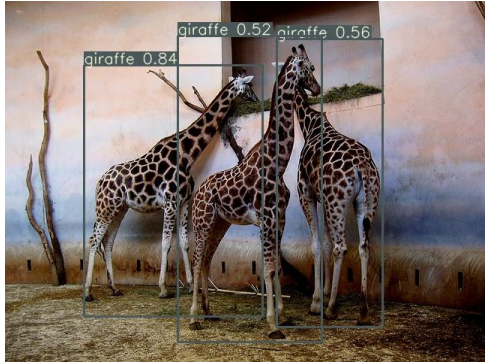
(c)
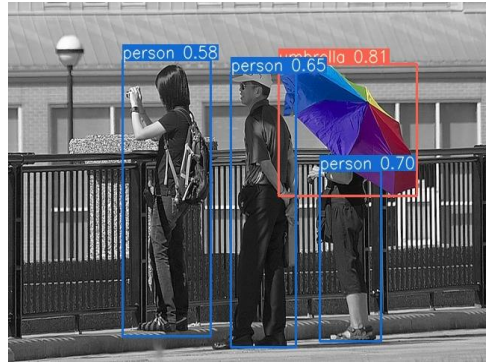
(d)
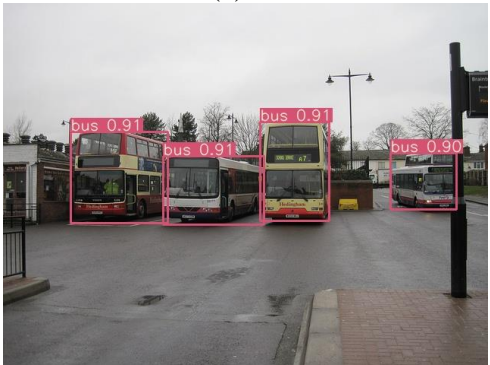
(e)
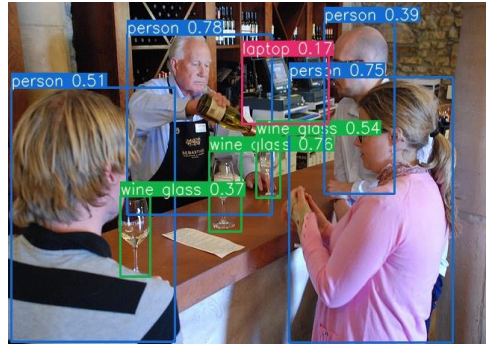
(f)

**FIGURE 6. Object detection results by YOLOv4-tiny.**

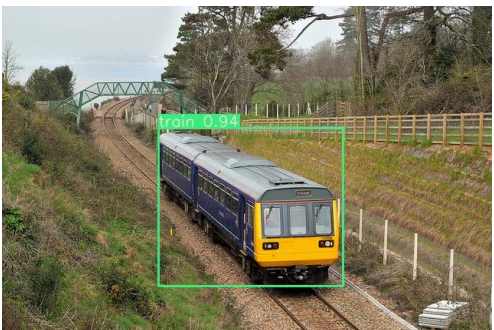
(a)
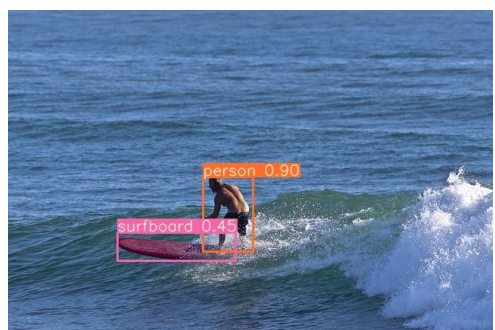
(b)

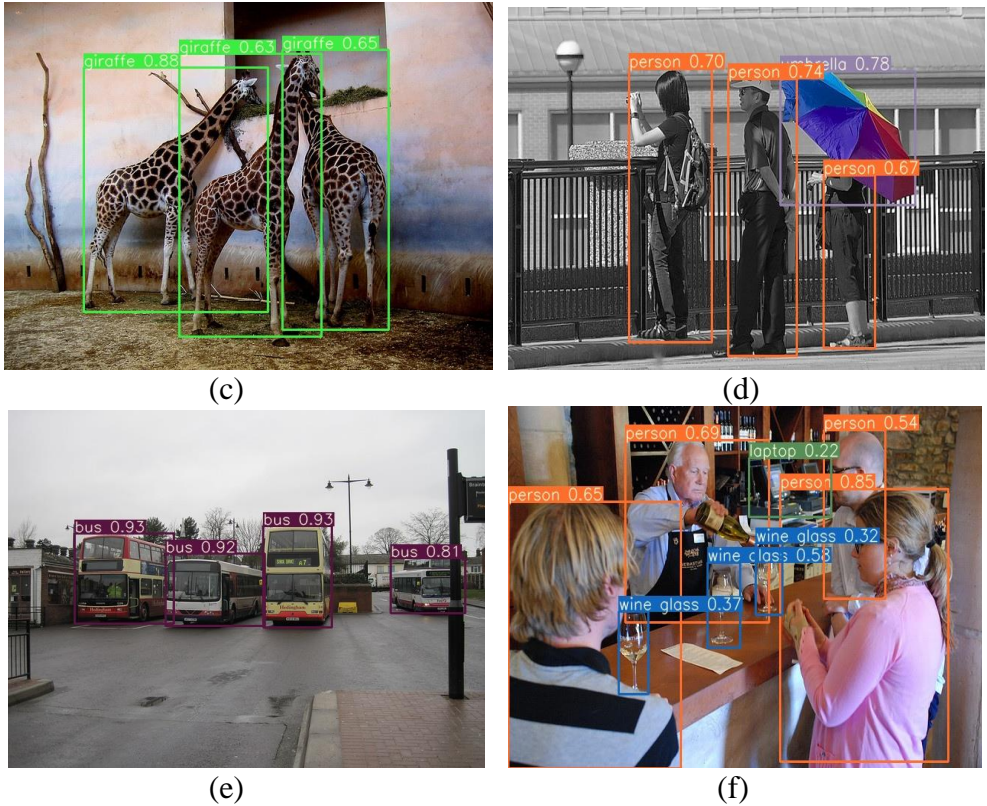

**FIGURE 7. Object detection results by our proposed method**

To test the performance of proposed method on difference devices, we simulate the different methods on CPU and Raspberry Pi that is an embedded device, respectively. The CPU model is Intel Xeon E5-2678 v3 with 2.5 GHZ main frequency, and the Raspberry Pi model is Raspberry Pi 3B with BC219M2835 processor. we also use the same MS COCO dataset as testset. The size of input images is 416×416 and the batch size is 16. On the CPU, we use whole MS COCO dataset to test different methods. Due to the limited of storage space in Raspberry Pi device, we randomly select 40 images from MS COCO dataset to test different methods on the Raspberry Pi. Figure 9 is the Raspberry Pi device that is testing different methods. The screenshot of simulation results that are obtained by Raspberry Pi device is shown in figure 10. In figure 10, the time used to recognize the 40 images for YOLOv3-tiny, YOLOv4-tiny and our proposed method are 219s, 211s and 128s, respectively. We also transform the consumed time to FPS by dividing the time by the number of recognized images. The FPS is shown in figure 10. In figure 10, the first image is the FPS obtained by using CPU to test different methods, and second image is the FPS obtained by using Raspberry Pi. On the CPU, the number of FPS for YOLOv3-tiny, YOLOv4-tiny and our proposed method are 32, 25 and 37, respectively. Compared with YOLOv3-tiny and YOLOv4-tiny, the FPS of our proposed method increases by 15% and 48%, respectively. On the Raspberry Pi, the number of FPS for YOLOv3-tiny, YOLOv4-tiny and our proposed method are 0.18, 0.19 and 0.31, respectively. Compared with YOLOv3-tiny and YOLOv4-tiny, the FPS of our proposed method increases by 72% and 63%, respectively. Based on the above analysis, the differences in FPS between our proposed method and YOLOv3-tiny and YOLOv4-tiny are larger on Raspberry Pi than on CPU. This means that our proposed method is more suitable for developing on embedded devices.

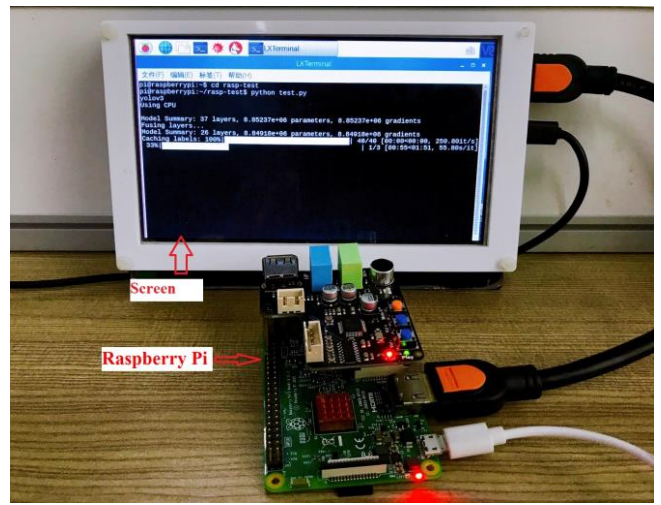

**FIGURE 8. Raspberry Pi device.**

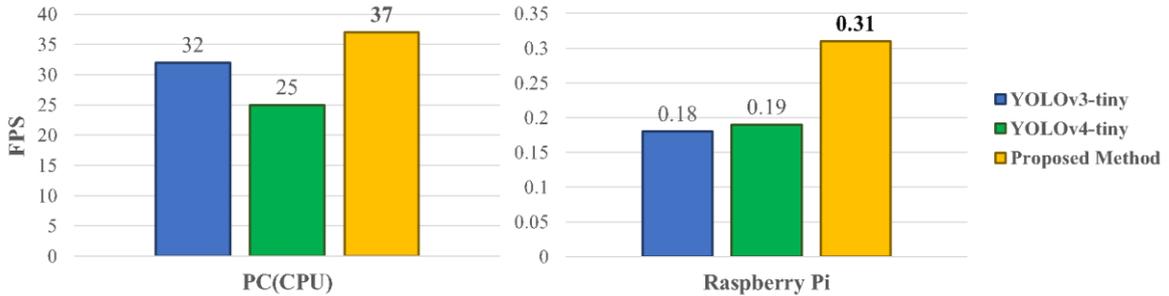

**FIGURE 9.** Simulation results obtained by Raspberry Pi.

**FIGURE 10.** The number of FPS for different methods on CPU and Raspberry Pi.

## V. CONCLUSION

This paper proposes an improved YOLOv4-tiny method in terms of network structure. To reduce the consuming time of object detection, we use two same ResBlock-D modules to replace two CSPBlock modules in YOLOv4-tiny network to simple the network structure. To balance the object detection time and accuracy, we design auxiliary network block by using two $3\times 3$ convolutions network, channel attention, spatial attention and concatenate operation to extract global feature. In the end, we merge our designed auxiliary network into the backbone network to build new backbone network. This realizes the convergence between deep network and shallow network. It makes the improved backbone network can extract the global and local feature of detection object, and further more improve the accuracy of detection without increasing large calculation. Compared with YOLOv3-tiny and YOLOv4-tiny, the proposed method has faster object detection speed and almost the same mean value of average precision with YOLOv4-tiny.